# Hybrid Processing of Beliefs and Constraints*

Rina Dechter and David Larkin

Department of Information and Computer Science
University of California, Irvine, CA 92697-3425
{dechter,dlarkin}@ics.uci.edu

## Abstract

This paper explores algorithms for processing probabilistic and deterministic information when the former is represented as a belief network and the latter as a set of boolean clauses. The motivating tasks are 1. evaluating belief networks having a large number of deterministic relationships and 2. evaluating probabilities of complex boolean queries or complex evidence information over a belief network. We present and analyze a variable elimination algorithm that exploits both types of information, and provide empirical evaluation demonstrating its computational benefits.

## 1 Introduction and motivation

The paper addresses the question of processing deterministic relationships that interact with probabilistic information expressed as belief networks. Two primary sources of determinism are considered: network-based and query-based. Network determinism means that a portion of the probabilistic network contains deterministic relationships, such as OR, AND and Parity functions. A second source of determinism can be generated outside the knowledge-base, when evaluating the posterior belief of complex constraint-based queries, or when given complex evidence structure (e.g., disjunctive information).

We will show that both sources of determinism can be reduced to evaluating the probability of Boolean queries. While we will assume that the deterministic information is expressed as boolean formulas in conjunctive normal form (CNF), the framework is extensible, in principle, to relational constraint expressions over multi-valued domains.

The paper presents a variable-elimination algorithm for computing the probability of a CNF query over a belief network. It is known that such queries can be handled by modeling the formula as part of the belief network ([Pearl, 1988]). However, as we demonstrate, it is computationally beneficial to distinguish between the deterministic and probabilistic information. It facilitates constraint processing, especially search and *constraint propagation* (e.g. unit resolution), which has proven essential for efficient processing of Boolean and constraint expressions. We analyze the algorithm's complexity based on its dependency graph. Preliminary experiments show that exploiting deterministic information can lead to significant speedup of up to a factor of 2 on the average.

## 2 Preliminaries and background

Let $X = \{X_1, ..., X_n\}$ be a set of random variables over multi-valued domains, $D_1, ..., D_n$, respectively. A *belief network* is a pair $(G, P)$ where $G = (X, E)$ is a directed acyclic graph over the variables, and $P = \{P_i\}$, where $P_i$ denotes conditional probability tables (CPTs) $P_i = \{P(X_i|pa_i)\}$, and $pa_i$ is the set of *parent* nodes pointing to $X_i$ in the graph. When the CPTs entries are "0" or "1" only, they are called *deterministic or functional CPTs*. When some of the CPT's entries are "0" or "1" they are called *mixed CPTs*. The family of $X_i$, $F_i$, includes $X_i$ and its parent variables. The belief network represents a probability distribution over $X$ having the product form $P(x_1, ..., x_n) = \Pi_{i=1}^{n} P(x_i|x_{pa_i})$ where an assignment $(X_1 = x_1, ..., X_n = x_n)$ is abbreviated to $x = (x_1, ..., x_n)$ and where $x_S$ denotes the restriction of a tuple $x$ over a subset of variables $S$. An evidence set $e$ is an instantiated subset of variables. We use upper case letters for variables and nodes in a graph and lower case letters for values in a variable's domain. The scope of an arbitrary function is its set of arguments. The *moral graph* of a directed graph is the undirected graph obtained by connecting the parent nodes of each variable and eliminating direction.

*This work was supported in part by NSF grant IIS-0086529 and by MURI ONR award N00014-00-1-0617

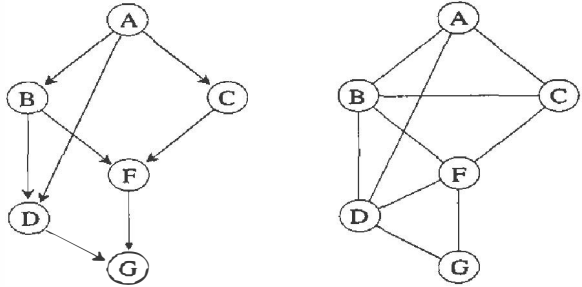

Figure 1: Belief network $P(g, f, d, c, b, a) = P(g|f,d)P(f|c,b)P(d|b,a)P(b|a)P(c|a)P(a)$

**Propositional theories.** Propositional variables which take only two values $\{true, false\}$ or $\{1, 0\}$, are denoted by uppercase letters $P, Q, R$. Propositional literals (i.e., $P, \neg P$) stand for $P = true$ or $P = false$, and disjunctions of literals, or *clauses*, are denoted by $\alpha, \beta, \ldots$. For instance, $\alpha = (P \vee Q \vee R)$ is a clause. A *unit clause* is a clause of size 1. The *resolution* operation over two clauses $(\alpha \vee Q)$ and $(\beta \vee \neg Q)$ results in a clause $(\alpha \vee \beta)$, thus eliminating $Q$. A formula $\varphi$ in conjunctive normal form (*CNF*) is a set of clauses $\varphi = \{\alpha_1, \ldots, \alpha_t\}$ that denotes their conjunction. The set of *models* or *solutions* of a formula $\varphi$, denoted $m(\varphi)$ is the set of all truth assignments to all its symbols that do not violate any clause. $resolve(\varphi, \alpha)$ is the set of resolvents of each clause in $\varphi$ with $\alpha$.

**Example 2.1** *Figure 1a gives an example of a belief network over 6 variables. Assume that the CPTs associated with C is mixed given by $P(C = 1|A = 0) = 1, P(C = 1, A = 1) = 0.5$ and that G is associated with a deterministic function: $G = D \vee F$. The rest of the CPTs are positive. The moral graph is given in Figure 1b.*

**Bucket elimination.** *Bucket elimination* is a unifying algorithmic framework for variable elimination algorithms applicable to probabilistic and deterministic reasoning [Bertele and Brioschi, 1972, N. L. Zhang and Poole, 1994, Dechter, 1996]. The input to a bucket-elimination algorithm is a set of functions or relations. Given a variable ordering, the algorithm partitions the functions (e.g., CPTs) into buckets, where a function is placed in the bucket of its latest argument in the ordering. The algorithm processes each bucket, from last to first, by a variable elimination procedure that computes a new function that is placed in an earlier (lower) bucket. The time and space complexity of such algorithms is exponential in a graph parameter called induced width $w^*$. For more information see [Dechter, 1999].

## 3 Tasks

The primary basic query over belief networks is *belief updating*, namely evaluating the posterior probability of each singleton proposition given some evidence. In this paper we address complex queries and complex evidence that are expressible as Boolean formulas on subsets of the variables. In addition we will discuss the processing of hybrid networks containing deterministic and mixed CPTs, and show that both explicit and implicit deterministic information in such networks can be exploited computationally by appropriate transformation to CNF query evaluation.

### 3.1 Complex queries, given complex evidence

**CNF Probability Evaluation (CPE).** The problem of evaluating the probability of CNF queries over belief networks has application to query answering in massive databases. In particular, for massive data archives, it is possible to construct an approximate model of the data *offline* using a belief network and then to answer real-time queries using the approximate model (without recourse to the original data) [Pavlov et al., 2000].

Another application is to network reliability. Given a communication graph with a source and destination, one seeks to diagnose failure of communication. Since several paths may be available, the reason for failure can be described by a CNF formula. Failure means that for all paths (conjunctions) there is a link on that path (disjunction) that fails. Given a probabilistic fault model of the network, the task is to assess the probability of a failure [Portinale and Bobbio, 1999].

DEFINITION **3.1 (CPE)**
*Given a belief network $(G, P)$, defined over propositional variables $X = \{X_1, ..., X_n\}$ and given a CNF query $\varphi$ over a subset $Q = \{Q_1, ...Q_r\}$, where $Q \subseteq X$, the CNF Probability Evaluation (CPE) is to find the probability $P(\varphi)$.*

**Complex evidence.** We can envision situations when one wants to assess belief of a proposition given partial, disjunctive information. For example, given that a customer purchased a coat or a shirt, but did not buy a tie, what is the probability that they will also purchase shoes? This type of query is very valuable for predictive modeling, e.g., "cross-sell" applications where we determine which other products a customer is likely to purchase.

*Belief assessment conditioned on a CNF evidence* is the task of assessing $P(X|\varphi)$ for every variable $X$. Since $P(X|\varphi) = \alpha P(X \wedge \varphi)$ when $\alpha$ is a normalizing constant relative to $X$, computing $P(X|\varphi)$ reduces to a CPE task for the query $((X = x) \wedge \varphi)$. More generally, $P(\varphi|\psi)$ can be derived from $P(\varphi|\psi) = \alpha_\varphi \cdot P(\varphi \wedge \psi)$ when $\alpha_\varphi$ is a normalization constant relative to all the models of $\varphi$.

A CNF query can also be defined over multi-valued variables $X_1, ...X_n$. Its propositions are $(X_i, a)$, where $a \in D_i$. The proposition is true if $X_i$ is assigned value $a \in D_i$ and is false otherwise. The CNF is augmented with a collection of 2-CNFs for each variable, that forbids assignments of more than one value to a variable. Namely, for every $i$ $(X_i, a) \rightarrow \neg(X_i, b)$ if $a \neq b$.

### 3.2 Evaluating beliefs in hybrid networks

Often belief networks have a hybrid probabilistic and deterministic relationships. Such networks appear in medical applications in coding networks [R.J. McEliece and Cheng, 1997] and in networks having CPTs that are *causally independent* [Heckerman, 1989]. Recent work in dynamic decision networks reveals the need to express large portion of the knowledge using deterministic constraints. We argue that treating such information in a special manner, using constraint processing methods is likely to yield significant computational benefit.

**Hybrid networks** A hybrid belief network (HBN) is a triplet $< G, P, F >$, $G = (X, E)$, where $X$ is a set of variables partitioned into $X = R \cup D$. Variables in $R$ are probabilistic and have regular CPTs while variables in D are deterministic having a function defined from their parents to the variable. The CPTs of probabilistic variables can be positive or mixed. In the latter case some probability entries in the CPTs are 0 or 1.

**Belief assessment in an HBN** translates to a CPE task. The idea is to collect together all the deterministic information appearing in the functions of $F$ and to extract the deterministic information in the mixed CPTs, and then transform it all to one CNF expression. This expression can then be treated as a CNF query over the original network. Clearly, every function can be expressed as a CNF formula. Also, each entry in a mixed CPT $P(X_i|pa_i)$, having $P(x_i|x_{pa_i}) = 1$, ($x$ is a tuple of variables in the family of $X_i$) can be translated to the clause $x_{pa_i} \rightarrow x_i$, and all such entries constitute a conjunction of clauses.

Let $HBN = < C, P, F >$ be a hybrid network. Given evidence $e$, assessing the posterior probability of a single variable $X$ given evidence $e$ is to compute $P(X|e) = \alpha P(X \wedge e)$. Let $cl(P)$ be the clauses extracted from the mixed CPTs, and let $cl(F)$ be the clauses expressing the conjunction of functions in $F$. The network's deterministic portion is $cl(F) \wedge cl(P)$, and because this conjunction is redundant relative to the given network, namely since $P(cl(F) \wedge cl(P)) = 1$ we can write:

$P((X = x) \wedge e) = P((X = x) \wedge e \wedge cl(F) \wedge cl(P))$
Therefore, to evaluate the belief of $X = x$ we can evaluate the probability of the CNF formula $\varphi = ((X = x) \wedge e \wedge cl(F) \wedge cl(P))$ over the original HBN. While some of the information is expressed redundantly, both in the network and in the query, it is semantically correct.

**Example 3.1** *Consider the HBN in Figure 1. We can extract the clauses $\varphi = \{(\neg D \vee G), (\neg F \vee G), (\neg G \vee D \vee F)\}$ from the only deterministic function $G = D \vee F$. From the mixed CPT of C we can extract the clause $(A \vee C)$. Answering the query $P(X \wedge \neg G)$ when $X$ is any variable is equivalent to evaluating $P(X \wedge \neg G, \wedge(\neg D \vee G) \wedge (\neg F \vee G) \wedge (\neg G \vee D \vee F) \wedge (A \vee C)\}$.*

## 4 Bucket-elimination for CPE

The following paragraphs derive a bucket-elimination algorithm for CPE. This is a straightforward extension of the variable elimination algorithm Elim-bel for belief updating [Dechter, 1996]. Given a belief network defined over variables $X = \{X_1, ..., X_n\}$ and a CNF query $\varphi$ over[1] $Q \subseteq X$, where the size of $Q$ is $r$, the $CPE$ task is to compute a sum of probabilities of all the models of $\varphi$, namely: $P(\varphi) = \sum_{\bar{x}_Q \in m(\varphi)} P(\bar{x}_Q)$ where $\bar{x} = (x_1, ..., x_n)$. Using the belief-network product form we get: $P(\varphi) = \sum_{\{\bar{x}|\bar{x}_Q \in m(\varphi)\}} \prod_{i=1}^n P(x_i|x_{pa_i})$. For derivation purpose, we next assume that $X_n$ is one of the query variables, and we separate the summation over $X_n$ and $X - \{X_n\}$. We denote by $\gamma_n$ the set of all clauses containing $X_n$ and by $\beta_n$ all the rest of the clauses. The scope of $\gamma_n$ is denoted $Q_n$, $S_n = X - \{X_n\}$ and $U_n$ is the set of all variables in the scopes of the CPTs and clauses that mention $X_n$. We define $\bar{x}_i = (x_1, ..., x_i)$. We get:

$$P(\varphi) = \sum_{\{\bar{x}_{n-1}|\bar{x}_{S_n} \in m(\beta_n)\}} \sum_{\{x_n|\bar{x}_{Q_n} \in m(\gamma_n)\}} \prod_{i=1}^n P(x_i|x_{pa_i})$$

Denoting by $t_n$ the indices of functions in the product that *do not* mention $X_n$ and by $l_n = \{1, ...n\} - t_n$ we get:

$$P(\varphi) = \sum_{\{\bar{x}_{n-1}|\bar{x}_{S_n} \in m(\beta_n)\}} \prod_{j \in t_n} P_j \cdot \sum_{\{x_n|\bar{x}_{Q_n} \in m(\gamma_n)\}} \prod_{j \in l_n} P_j$$

Therefore:

$$P(\varphi) = \sum_{\{\bar{x}_{n-1}|\bar{x}_{S_n} \in m(\beta_n)\}} (\prod_{j \in t_n} P_j) \cdot \lambda^{X_n} \quad (1)$$

---
[1] It is easy to extend this to propositions over multi-valued variables

where $\lambda^{X_n}$ over $U_n - \{X_n\}$, is defined by

$$\lambda^{X_n} = \sum_{\{x_n | \bar{x}_{Q_n} \in m(\gamma_n)\}} \prod_{j \in l_n} P_j \qquad (2)$$

Therefore, if we place all CPTs and clauses mentioning $X_n$ into the bucket of $X_n$ we can compute the function in EQ. ( 2). The computation of the rest of the expression proceeds with $X_{n-1}$, using EQ. (1), in the same manner.

**Case of observed variables.** When $X_n$ is observed, or constrained by a literal, the summation operation reduces to assigning the observed value to each of its CPTs *and* to each of the relevant clauses. In this case EQ. (2) becomes (assume $X_n = x_n$ and $P_{=x_n}$ is the function instantiated by assigning $x_n$ to $X_n$):

$$\lambda^{x_n} = \prod_{j \in l_n} P_{j=x_n}, \quad if\ \bar{x}_{Q_n} \in m(\gamma_n \wedge (X_n = x_n)) \quad (3)$$

Otherwise, $\lambda^{x_n} = 0$. Since $\bar{x}_{Q_n}$ satisfies $\gamma_n \wedge (X_n = x_n)$ only if $\bar{x}_{Q_n - X_n}$ satisfies $\gamma^{x_n} = resolve(\gamma_n, (X_n = x_n))$, we get:

$$\lambda^{x_n} = \prod_{j \in l_n} P_{j=x_n} \quad if\ \bar{x}_{Q_n - X_n} \in m(\gamma_n^{x_n}) \qquad (4)$$

Therefore, we can extend the case of observed variable in a natural way: CPTs are assigned the observed value as usual while clauses are individually resolved with the unit clause $(X_n = x_n)$, and both are moved to appropriate lower buckets.

Algorithm Elim-CPE, described in Figure 2, includes therefore a limited amount of constraint propagation in the form of unit-resolution. Thus, for the variable ordering of choice, once all CPTs and clauses are partitioned (each clause and CPT is placed in the latest bucket of its scope), we process the buckets from last to first. If the bucket contains a literal we assign its value to the CPTs, resolve it with the clauses and move the resulting functions and clauses to the appropriate bucket. Otherwise, in each bucket we generate the probabilistic function. From our derivation it follows that

**THEOREM 4.1 (Correctness and Completeness)**
*Algorithm Elim-CPE is sound and complete for the CPE task.* □

Note that the algorithm includes also a dynamic reordering of the buckets that prefers processing buckets that include unit clauses. This may have a significant impact on efficiency because observations (namely unit clauses) avoid the creation of new dependencies.

**Example 4.2** *Lets treat the belief network in Figure 1 as if all its CPTs are pure positive, and assume we*

---

**Algorithm Elim-CPE**
**Input:** A belief network $(G, P)$, $P = \{P_1, ..., P_n\}$; A CNF formula on $r$ propositions $\varphi = \{\alpha_1, ... \alpha_m\}$ an ordering, $d$
**Output:** The belief $P(\varphi)$.
1. **Initialize:** Place buckets with unit clauses last in the ordering (to be processed first). Partition $P$ and $\varphi$ into $bucket_1$, ..., $bucket_n$, in the usual manner. (We denote probabilistic functions as $\lambda$s and clauses by $\alpha$s). Scopes of CPTs are denoted by $S$, of clauses by $Q$.
2. **Backward:** Process from last to first.
Let $P$ be the current bucket.
For $\lambda_1, ..., \lambda_j$, $\alpha_1, ..., \alpha_r$ in $bucket_p$, do
• If $bucket_p$ contains $X_p = x_p$ (or a unit clause),
a. Assign $X_p = x_p$ to each $\lambda_i$
b. **Resolve** each $\alpha_i$ with the unit clause, and put resolvents and probabilistic function lower buckets and
c. Move any bucket with unit clause to top of processing.
• **Else, compute probabilistic function** $\lambda^P = \sum_{\{x_p | \bar{x}_{U_p} \in m(\alpha_1, ..., \alpha_r)\}} \prod_{i=1}^{j} \lambda_i$,
over $U_p = S \cup Q - \{X_p\}$, $S = \cup_i S_i$, $Q = \cup_j Q_j$, and place any generated function or clause into its appropriate lower bucket.
3. **Return** $P(\varphi)$ generated in the first bucket.

Figure 2: Algorithm Elim-CPE

*get the query $\varphi = (B \vee C) \wedge (G \vee D) \wedge (\neg D \vee \neg B)$. The initial partitioning into buckets along the ordering $d = A, C, B, D, F, G$, as well as the output buckets are given in Figure 3a. In bucket G we compute: $\lambda^G(f, d) = \sum_{\{g | g \vee d = true\}} P(g|f, d)$. In bucket F: $\lambda^F(b, c, d) = \sum_f P(f|b, c) \lambda^G(f, d)$. In bucket D: $\lambda^D(a, b, c) = \sum_{\{d | \neg d \vee \neg b = true\}} P(d|a, b) \lambda^F(b, c, d)$. In bucket B: $\lambda^B(a, c) = \sum_{\{b | b \vee c = true\}} P(b|a) \lambda^D(a, b, c)$. In bucket C: $\lambda^C(a) = \sum_c P(c|a) \lambda^B(a, c)$. In bucket A: $\lambda^A = \sum_a P(a) \lambda^C(a)$  $P(\varphi) = \lambda^A$.*

*Let's now extend the example by adding $\neg G$ to the query. This will place $\neg G$ in the bucket of G (See Figure 3b.) The Figure shows the derived functions and clauses, demonstrating the effect of unit resolution. Note the change in bucket ordering due to the preference to processing buckets with unit clauses.*

The following example extract clauses from the CPTs and then applies Elim-CPE.

**Example 4.3** *Consider again the belief network in Figure 1 and the query $P(A|\neg G)$ but assume the deterministic and mixed CPTs as described in Example 3.1. The extracted CNF is $\varphi = (\neg D \vee G) \wedge (\neg F \vee G) \wedge (\neg G \vee D \vee F) \wedge (A \vee C)$. The initial partitioning into buckets along the ordering $d = A, B, C, D, F, G$, as well as the output buckets are given in Figure 4a. In bucket G, since we have a unit clause, we compute: $\lambda^G(f, d) = P(G = 0|D, F)$. Applying unit resolution yields the literals $\neg F$ and $\neg D$. Since we have* a unit clause in bucket $F$, it will be assigned, yielding $\lambda^F(b,c) = P(F = 0|b,c)$. In bucket $D$ we have a generated unit clause $\neg D$ causing an assignment: $\lambda^D(a,b) = P(d = 0|a,b)$ and $\lambda^D = \lambda^F(D = 0)$. In bucket $C$: $\lambda^C(a,b) = \sum_{\{b|a\vee c=true\}} P(c|a)\lambda^F(b,c)$. Since the clause $A \vee C$ was extracted from $P(C|A)$ there is a redundancy in the above computation. Instead we will generate the function $\lambda^C(a,b) = \sum_b P(c|a)\lambda^F(b,c)$ which may save time, depending on the implementation. In bucket $B$: $\lambda^B(a) = \sum_c P(b|a)\lambda^C(a,b)\lambda^D(a,b)$. In bucket $A$: $\lambda^A(a) = P(a)\lambda^B(a)\lambda^D$. $P(A|\neg G) = \alpha\lambda^A(a)$. Regular Elim-CPE, not extracting deterministic CNF information, creates functions on 3 variables as is shown in Figure 4b.

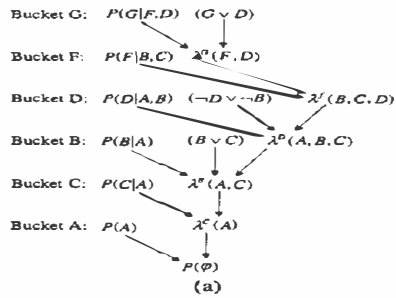
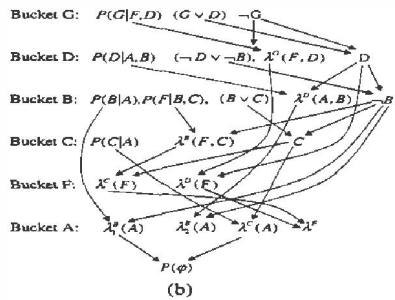

Figure 3: Trace of Elim-CPE (a) no observation (b) with observation

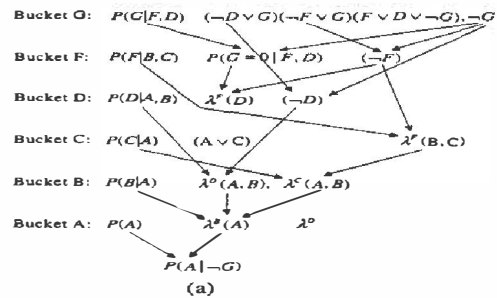
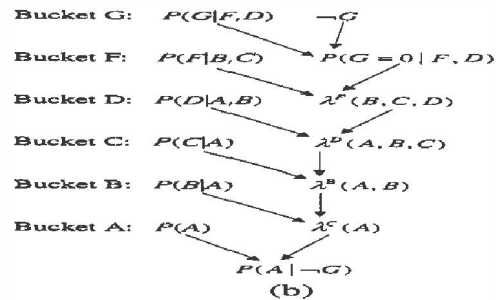

Figure 4: Variable elimination for a hybrid network: (a) Elim-CPE with clause extraction (b) regular Elim-CPE

Algorithm Elim-CPE-D is geared towards processing hybrid networks. It first extracts deterministic clauses from deterministic CPTs, and then applies Elim-CPE. However, for efficiency's sake, the new clauses are used for resolutions only in each bucket and are ignored for function computation.

### 4.1 Complexity

**Induced-graphs and induced width.** The *width of a node* in an ordered graph is the number of the node's neighbors that precede it in the ordering. The *width of an ordering d*, denoted $w(d)$, is the maximum width over all nodes. The *induced width of an ordered graph*, $w^*(d)$, is the width of the induced ordered graph obtained as follows: nodes are processed from last to first; when node $X$ is processed, all its preceding neighbors are connected. The *induced width of a graph*, $w*$, is the minimal induced width over all its orderings [Arnborg, 1985].

As usual, the complexity of bucket elimination algorithms is related to the number of variables appearing in each bucket. The worst-case complexity is time and space exponential in the size of the maximal bucket, which is captured by the induced-width of the relevant graph. Given a belief network and a query $\varphi$, the *augmented graph* of the network is the moral graph with additional arcs between each two variables appearing in the same clause of the CNF.

Consider now the computation inside a bucket. If $\gamma_P$ is the CNF theory in bucket $P$, defined over subset $Q_p$, and $\lambda_1,....\lambda_j$ are the probability functions whose union of scopes is $S_p$, we compute: $\lambda^P = \sum_{\{x_p|\bar{x}_Q \in m(\gamma_P)\}} \prod_i \lambda_i$ whose scope is $U_p = Q_p \cup S_p - \{X_p\}$. A brute force computation of this expression is $O(exp(|U_p| + 1))$. Since $|U_p|$ is bounded by $w^*(d)$ of the augmented graph, along $d$, the complexity of Elim-CPE is $O(n \cdot exp(w^*(d)))$.

To capture the simplification associated with observed variables or unit clauses, we connect only parents of each non-observed variable when generating the induced graph. The *adjusted induced width* is the width of this adjusted induced-graph. For details see [Dechter and Larkin, 2001]. In summary,

THEOREM 4.4 *Given a CNF $\varphi$ and an ordering $o$, the complexity of Elim-CPE is time and space $O(n \cdot exp(w^*(o)))$, where $w^*(o)$ is the induced width along $o$ of the augmented graph adjusted relative to the observed variables and unit clauses generated by unit-resolution, in $\varphi$.* □

### 4.2 Bucket-elimination with hidden variables

Consider now the alternative of modeling clauses as CPTs. It requires expressing each clause as a CPT with a new hidden variable and the addition of evidence to the hidden nodes. Subsequently we can apply a regular variable elimination algorithm ([Dechter, 1996, N. L. Zhang and Poole, 1994]). We call the resulting algorithm Elim-Hidden.

There is no substantial difference between Elim-CPE and Elim-Hidden in terms of worst-case complexity. Processing the hidden variables creates tables that corresponds to the clauses which are placed in the same buckets that the original clauses occupy in Elim-CPE; producing just a linear overhead. Subsequently, when computing the function's bucket, Elim-Hidden uses multiplication to factor out non-models and Elim-CPE uses summation over models. In example 4.3, Elim-Hidden is far inferior, unable to recognize unit clauses.

### 4.3 Elim-CPE with constraint propagation

Constraint propagation can, in principle, improve Elim-CPE by inferring new unit clauses beyond the power of unit-resolution. Furthermore, inferred clauses correspond to infered conditional probabilities that are either "0" or "1".

One form of constraint propagation is bounded resolution [Rish and Dechter, 2000]. It applies pair-wise resolution to any two clauses in the CNF theory iff the resolvent does not exceed a bounding parameter, $i$. Bounded-resolution algorithms can be applied until quiesence or in a directional manner, called $BDR(i)$. After partitioning the clauses into ordered buckets, each is processed by resolution with bound $i$.

We extend Elim-CPE into a parameterized family of algorithms Elim-CPE(i) that incorporates $BDR(i)$. The added operation in $bucket_p$ is: (If the bucket does not have an observed variable)
For each pair $\{(\alpha \vee Q_i), (\beta \vee \neg Q_i)\} \subseteq bucket_i$. If the resolvent $\gamma = \alpha \cup \beta$ contains no more than $i$ propositions, place the resolvents in the bucket of its highest index variable. Higher levels of propagation may infer more unit-clauses and general nogoods but require more computation. It is hard to assess in advance the right balance of constraint propagation. It is known that the complexity of $BDR(i)$ is $O(exp(i))$. Therefore, for small levels of $i$ the computation in non-unit

| Algorithm | Time | mf | C. | U. |
|---|---|---|---|---|
| Elim-CPE: | 18 | 18 | 18 | 2 |
| Elim-Hidden: | 33 | 19 | 0 | 0 |

Figure 5: 50 test instances, network parameters of $< 50, 5, 0 >$ and query parameters $< 50, 15 >$

| Algorithm | Time | mf | C. | U. |
|---|---|---|---|---|
| Elim-CPE: | 5 | 16 | 22 | 3 |
| Elim-Hidden: | 18 | 18 | 0 | 0 |

Figure 6: Averages over 35 test instances, network parameters of $< 40, 5, 0 >$ and query parameters $< 60, 10 >$

buckets is likely to be dominated by generating the probabilistic function rather than by $BDR(i)$.

## 5 Empirical Evaluation

There were four algorithms to be compared empirically: Elim-CPE (which is the same as Elim-CPE(0)), Elim-CPE(i), Elim-Hidden, and Elim-CPE-D. Some random networks were tested, as well as two realistic networks, the hailfinder and insurance networks. We report only some of the results for space reasons. For more information see [Dechter and Larkin, 2001].

**The random generator**. The test generator is divided into two parts. The first creates a random belief network using a tuple $< n, f, d >$ as a parameter, where $n$ is the number of variables, $f$ is the maximum family size, and $d$ is the fraction of deterministic entries in CPTs. Parents are chosen at random from the preceding variables in a fixed ordering. The entries of the CPT's are filled in randomly. The second part generates a 3-CNF query, using a pair of parameters $< c, e >$ where $c$ is the number of 3-CNF clauses (clauses are randomly chosen and each is given a random truth value) and $e$ is the number of observations.

All algorithms use min-degree order, computed by repeatedly removing the node with the lowest degree from the graph and connecting all its neighbors.

**Results on Random networks.**
*Elim-CPE vs Elim-Hidden.* We report first some of our results on Elim-CPE vs Elim-Hidden with two sets of random networks generated with parameters $< 50, 5, 0 >$ and $< 40, 4, 0 >$. The results of those runs are summarized in Figures 5 and 6 respectively. In the tables, the time is given in seconds, $C$ stands for derived Clauses, $U$ stands for derived Unit clauses, and $mf$ is the arity of the largest function created by the algorithm. Clearly $mf \leq w^*$.

We see that Elim-CPE-Hidden is slower than Elim-

| Algorithm | Time | mf | C. | U. |
|---|---|---|---|---|
| Elim-CPE(n): | 22 | 17 | 23 | 2 |
| Elim-CPE(3): | 21 | 17 | 20 | 2 |
| Elim-CPE(2): | 20 | 17 | 17 | 2 |
| Elim-CPE(1): | 18 | 17 | 15 | 2 |

Figure 7: Averages over 30 test instances with network parameters of $< 50, 5, 0 >$ and query parameters $< 50, 15 >$

| Algorithm | O. | Time | mf | C. | U. | F. |
|---|---|---|---|---|---|---|
| Elim-CPE-D: | 10 | 32 | 8 | 299 | 3 | 351 |
| Elim-CPE(0): | 10 | 60 | 16 | 0 | 0 | 0 |
| Elim-CPE-D: | 15 | 10 | 7 | 272 | 3 | 350 |
| Elim-CPE(0): | 15 | 33 | 15 | 0 | 0 | 0 |

Figure 8: Averages of 50 instances with network parameters $< 80, 4, 0.75 >$ and varied number of evidence

CPE by a factor of 2 on the average. This is expected because of Elim-CPE's constraint propagation, which creates more unit variables.

*Testing Elim-CPE(i).* The purpose in testing Elim-CPE(i) was to evaluate the effect of different levels of bounded i-resolution. Higher values of i may produce more clauses, especially unit clauses, which should speed up the computation. We ran the algorithm on networks generated by parameters of $< 50, 5, 0 >$ and with query parameters $< 50, 15 >$. The results are summarized in Figure 7. As we see in these tests, higher levels of constraint propagation were not successful in creating more unit clauses. It appears that larger and harder CNF queries are necessary to make stronger constraint propagation cost-effective.

*Testing Elim-CPE-D.* Figure 8 shows some tests of Elim-CPE-D vs. Elim-CPE on random networks. The difference is that Elim-CPE-D extracts deterministic information from CPT's. O. stands for the number of observed variables and F. stands for the number of clauses extracted from CPT's. We see that Elim-CPE-D was generally superior. The results for 10 unit clauses are also shown in the scatter diagram in Figure 9.

### Realistic Benchmarks

*Tests on Insurance network.* Next we tested the insurance network which is a realistic network for evaluating car insurance risks that contains deterministic information. It has 27 variables. In the experiments reported in Figure 10, Elim-CPE-D outperformed Elim-CPE substantially. Figure 12 contrasts Elim-CPE with Elim-Hidden on the insurance network.

*Testing on Hailfinder network.* Finally we tested the hailfinder network, another benchmark network that has 56 variables and includes deterministic information. It is a normative system that forecasts severe summer hail in northeast Colorado. The results are reported in Figure 11. Here again the results are consistent with earlier observations that Elim-CPE-D was the most efficient.

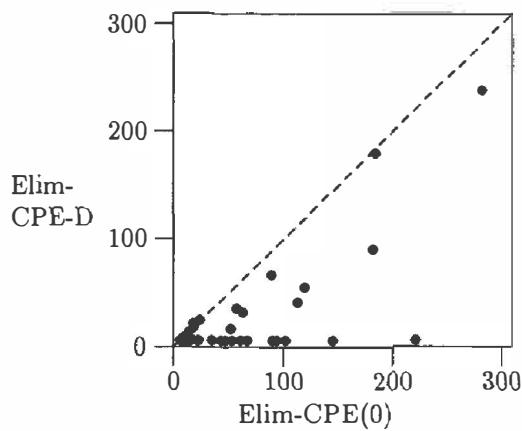

Figure 9: 48 test instances with network parameters $< 80, 4, 0.75 >$ and query parameters $< 0, 10 >$

| Algorithm | Time | mf | C. | U. | F. |
|---|---|---|---|---|---|
| Elim-CPE-D: | 48 | 8 | 210 | 1 | 302 |
| Elim-CPE(15): | 64 | 9 | 12 | 1 | 0 |
| Elim-CPE(0): | 61 | 9 | 6 | 0 | 0 |
| Elim-Hidden: | 104 | 10 | 0 | 0 | 0 |

Figure 10: 50 test instances of the insurance network (27 variables), with query parameters $< 20, 5 >$

| Algorithm | Time | mf | C. | U. | F. |
|---|---|---|---|---|---|
| Elim-CPE-D: | 4 | 4 | 269 | 1 | 501 |
| Elim-CPE(15): | 16 | 6 | 7 | 1 | 0 |
| Elim-CPE(0): | 16 | 6 | 7 | 1 | 0 |
| Elim-Hidden: | 33 | 7 | 0 | 0 | 0 |

Figure 11: 50 test instances of the hailfinder network (56 variables) with query parameters $< 15, 15 >$

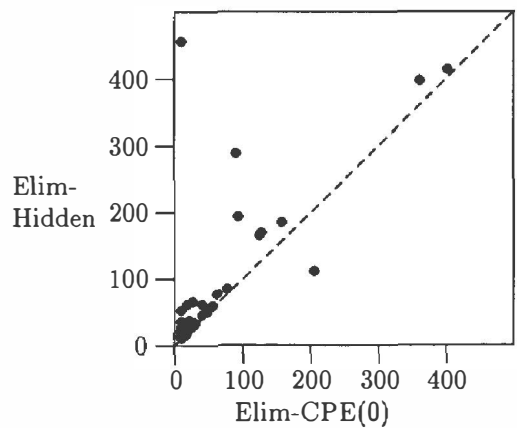

Figure 12: 50 test instances of the insurance network with query parameters $< 15, 5 >$

## 6 Discussion and related work

The most relevant work is that of Poole [Poole, 1997] providing a rule-based description of the conditional probability tables, and a variable elimination algorithm for exploiting this rule-based representation. When the information is deterministic, those rules are simple clauses, and their processing may reduce to simple resolution. I An area that uses heavily both deterministic and probabilistic information is planning under uncertainty. Most relevant is a recent stochastic planner called MAXPLAN [Majercik and Littman, ] which shows how stochastic planning can be transformed into an MAJSAT description and then solved by a search-based conditioning algorithm. It would be interesting to exploit our algorithm in the context of these works.

The paper presents a variable elimination algorithm called Elim-CPE, for answering Boolean CNF queries over a belief network. The algorithm is applicable to hybrid belief networks and to belief updating given partial information.

The nice property of the bucket-elimination algorithms is that their complexity is not dependent on the number of models in the CNF formula. Clearly, all the tasks addressed here could also be solved by conditioning search or by some combination of search and inference, and should be explored further. They avoid the space complexity of bucket elimination and may work well in practice.

The empirical results demonstrated that the proposed algorithm Elim-CPE, is far more effective than a brute force embedding of the CNF query into the belief network (i.e., Elim-Hidden) by a factor of 2 on the average, depending on the size of the CNF formula. When applying a variant of this algortihm to hybrid networks (i.e., Elim-CPE-D) we observed impressive improvement that were more significant as the portion of the deterministic information increased. Those results were consistent for randomly generated networks and some real benchmarks.

## References

[Arnborg, 1985] S. A. Arnborg. Efficient algorithms for combinatorial problems on graphs with bounded decomposability - a survey. *BIT*, 25:2–23, 1985.

[Bertele and Brioschi, 1972] U. Bertele and F. Brioschi. *Nonserial Dynamic Programming*. Academic Press, 1972.

[Dechter and Larkin, 2001] R. Dechter and D. Larkin. Hybrid processing of belief and constraints. *UCI Technical report, www.ics.uci.edu/ dechter*, 2001.

[Dechter, 1996] R. Dechter. Bucket elimination: A unifying framework for probabilistic inference algorithms. In *Uncertainty in Artificial Intelligence (UAI'96)*, pages 211–219, 1996.

[Dechter, 1999] R. Dechter. Bucket elimination: A unifying framework for reasoning. *Artificial Intelligence*, 113:41–85, 1999.

[Heckerman, 1989] D. Heckerman. A tractable inference algorithm for diagnosing multiple diseases. In *Uncertainty in Artificial Intelligence (UAI'89)*, pages 171–181, 1989.

[Majercik and Littman, ] S. M. Majercik and M. L. Littman. Maxplan: A new approach to probabilistic planning.

[N. L. Zhang and Poole, 1994] R. Qi N. L. Zhang and D. Poole. A computational theory of decision networks. *International Journal of Approximate Reasoning*, pages 83–158, 1994.

[Pavlov et al., 2000] D. Pavlov, H. Mannila, and P. Smyth. Probabilistic models for query approximation with 20 large sparse binary data sets. In *Submitted to UAI2000*, 2000.

[Pearl, 1988] J. Pearl. *Probabilistic Reasoning in Intelligent Systems*. Morgan Kaufmann, 1988.

[Poole, 1997] D. Poole. Probabilistic partial evaluation: Exploiting structure in probabilistic inference. In *IJCAI-97: Proceedings of the Fifteenth International Joint Conference on Artificial Intelligence*, 1997.

[Portinale and Bobbio, 1999] L. Portinale and A. Bobbio. Bayesian networks for dependency analysis: an application to digital control. In *Proceedings of the 15th Conference on Uncertainty in Artificial Intelligence (UAI99)*, pages 551–558, 1999.

[Rish and Dechter, 2000] I. Rish and R. Dechter. Resolution vs. search; two strategies for sat. *Journal of Automated Reasoning*, 24(1/2):225–275, 2000.

[R.J. McEliece and Cheng, 1997] D.J.C. MacKay R.J. McEliece and J.-F. Cheng. Turbo decoding as an instance of pearl's belief propagation algorithm. *IEEE J. Selected Areas in Communication*, 1997.